\documentclass{article}
\usepackage{spconf,amsmath,graphicx}
\usepackage{multirow}
\usepackage{booktabs}
\usepackage{placeins}
\usepackage{mathtools}
\usepackage{amsfonts}
\usepackage{comment}
\usepackage{enumitem}
\usepackage[flushleft]{threeparttable}
\title{Multi-task RNN-T with Semantic Decoder for Streamable Spoken Language Understanding}
\name{\begin{tabular}{c}Xuandi Fu$^{\dagger}$, Feng-Ju Chang, Martin Radfar, Kai Wei, \\ Jing Liu, Grant P. Strimel, Kanthashree Mysore Sathyendra\end{tabular}}

\address{Carnegie Mellon University$^{\dagger}$, Amazon Alexa \\ 
{\small \tt xuandifu@cmu.edu$^{\dagger}$, \{fengjc,radfarmr,kaiwe,jlmk,gsstrime,ksathyen\}@amazon.com}}

\begin{document}
\ninept
\maketitle
\begin{abstract}
End-to-end Spoken Language Understanding (E2E SLU) has attracted increasing interest due to its advantages of joint optimization and low latency when compared to traditionally cascaded pipelines.
Existing E2E SLU models usually follow a two-stage configuration where an Automatic Speech Recognition (ASR) network first predicts a transcript which is then passed to a Natural Language Understanding (NLU) module through an interface to infer semantic labels, such as intent and slot tags.
This design, however, does not consider the NLU posterior while making transcript predictions, nor correct the NLU prediction error immediately by considering the previously predicted word-pieces.
In addition, the NLU model in the two-stage system is not streamable, as it must wait for the audio segments to complete processing, which ultimately impacts the latency of the SLU system.
In this work, we propose a streamable multi-task semantic transducer model to address these considerations.
Our proposed architecture predicts ASR and NLU labels auto-regressively and uses a \emph{semantic decoder} to ingest both previously predicted word-pieces and slot tags while aggregating them through a fusion network. Using an industry scale SLU and a public FSC dataset, we show the proposed model outperforms the two-stage E2E SLU model for both ASR and NLU metrics.
\end{abstract}
\section{Introduction}
With the widespread application of intelligent voice assistants, e.g. Alexa, Siri, and Google Home, SLU systems have generated increased interest in the recent years.
An SLU system predicts semantic information implied by an audio signal.
This semantic content is commonly represented as intent, slot tags, named entities and/or part-of-speech taggings.
Today's SLU technology typically accomplishes this task in two separate stages, which we refer to as ASR-NLU approaches for SLU~\cite{larson2012spoken}: an ASR system first transcribes the audio signals~\cite{deepspeech, las}, and then the transcripts are passed to an NLU system to extract corresponding intent and slot tags~\cite{nlu, nlu2, bertnlu}; an example is presented in Table~\ref{tab:semantc_labels}.
Given the extracted semantic labels, downstream applications of the voice assistant can produce an appropriate response to the user.

Recently, complete E2E-SLU based approaches have attracted attention due to their efficiency and reduced model complexity compared with an ASR-NLU pipeline, making them suitable candidates for deployment on low-resource devices~\cite{google, lai2021semi, chuang2019speechbert, radfar2021fans}. 
Furthermore, performance improvements in both tasks, driven by joint training, has been observed in several studies \cite{google, fluent}.

\begin{figure*}[t]
\centering
\includegraphics[width=0.7\linewidth]{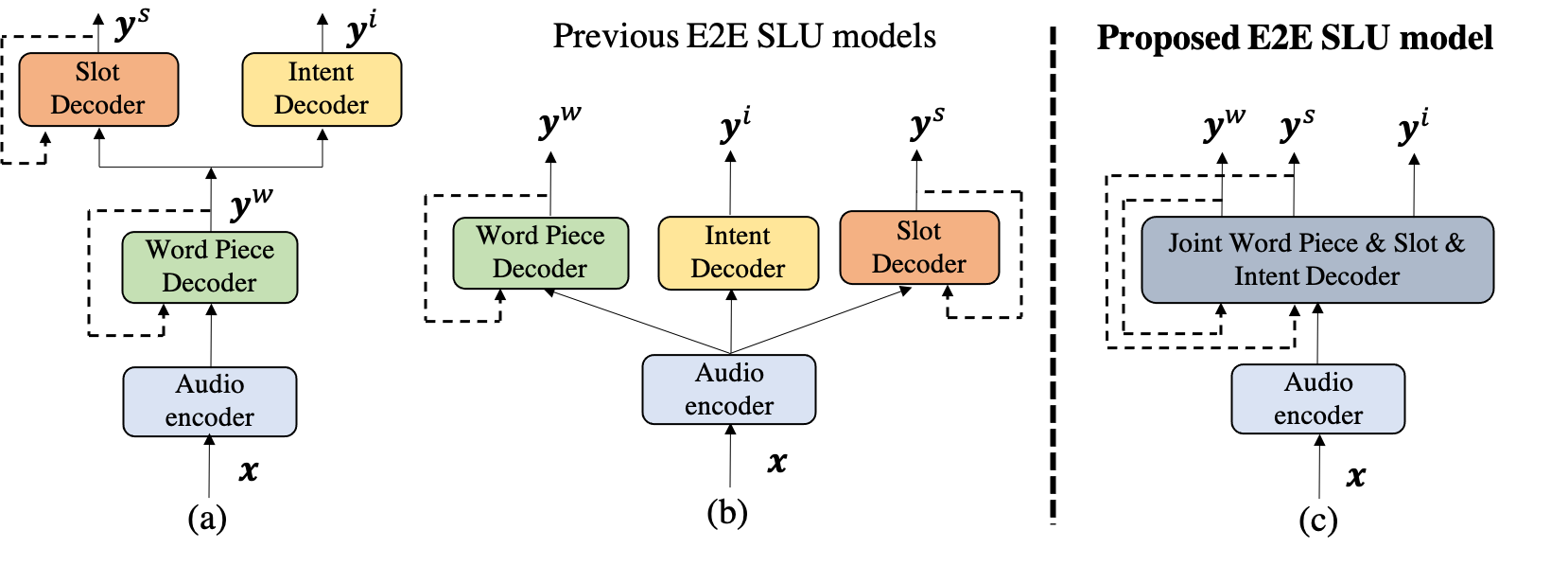}
\caption{A high-level diagram of comparing the proposed E2E SLU model to the previous two-stage~\cite{s2i,nlu, nlu2, bertnlu,larson2012spoken} (a) and one-stage \cite{google, radfar2021fans} (b) E2E SLU models. Dotted lines represent the conditioning of output label on its history, as a part of decoders.} 
\label{fig:high_level}
\vspace{-2.5mm}
\end{figure*}

Most of existing E2E-SLU models still adopt a two-stage setup as shown in Fig.~\ref{fig:high_level}(a), where the NLU subsystem waits for the transcripts of the whole utterance produced by the ASR subsystem to generate semantic labels~\cite{e2eslu, google, fluent, semantic1}. Meanwhile, the NLU subsystem is typically non-streamable. 
One-stage approaches to E2E-SLU have been proposed as well \cite{google, radfar2021fans}; however, again the NLU subsystem remains non-streamable. 
For example, in \cite{radfar2021fans}, slot tag prediction only occurs after the intent is extracted at the end of an utterance.
Moreover, in all the above approaches, the ASR and NLU label generations do not interact with one another during the forward pass of inference (Fig.~\ref{fig:high_level}(b)). 
As a result, this design can lead to three main limitations. 
First, the NLU posterior or hypothesis does not provide any feedback upon word-piece generation, while its feedback could be helpful to narrow down potential word-piece candidates generated in the next time step.
Secondly,  the decoding of the NLU label predictions is not streamable \cite{google}, given that the model is an encoder-decoder framework augmented by attention.
Finally, the inference speed of an SLU system may be affected by the nature of the cascaded setup and non-streamable NLU subsystem, all the while low latency is crucial for a responsive virtual assistant. 

To address these limitations, we propose a streamable E2E-SLU model based on RNN-T \cite{rnnt, rnnt-asr} with a novel semantic beam search decoder which predicts word-pieces and NLU labels jointly, as illustrated in Fig.~\ref{fig:high_level}(c).
Specifically, we introduce a semantic decoder to aggregate not only the word-pieces but also slot candidates during the beam search, which we call \emph{semantic beam search}. 
Furthermore, we propose different multi-task loss functions to learn the alignment between word-pieces and slot tags along with the intent prediction.

% An example of semantic labels
\begin{table}[t]
\centering
\small
\tabcolsep=0.1cm
\caption{\small An example of a transcription, slot tags, and intent.}
\label{tab:semantc_labels}
\begin{tabular}{cc}
\toprule
\textbf{transcription} & turn on the kitchen light \\ \midrule
\textbf{slot tags} & [DeviceLocation]: kitchen \\
& [ApplianceType]: light, [Other]:turn,on,the   \\ \midrule         
\textbf{intent} & TurnOnApplianceIntent \\ \bottomrule
\end{tabular}
\vskip -10pt
\end{table}

\section{Related Work}

A multi-task learning framework for E2E SLU was first introduced in \cite{google}; the authors investigated several encoder-decoder structures for joint training of ASR and NLU tasks, in which the multi-task structure achieves the best performance. 
This work was followed by proposed pre-training based approaches to improve the performance \cite{fluent, pretrain1, pretrain2}; all of these models are designed specifically for intent classification. 
Another category of work attempts a combined parameter transfer from well trained end-to-end ASR systems and end-to-end NLU models such as pretrained BERT~\cite{devlin2018bert} through teacher-student learning~\cite{semantic1, semantic2, semantic3}.
Note that both of these categories rely on at least a two-stage training process and all operate with non-streamable inference. 

Leveraging RNN-T, in \cite{s2i}, a two-stage E2E SLU structure was proposed where the RNN-T based ASR subsystem interacts with an NLU subsystem through an interface, which is not streamable.

In the most recent work, \cite{streaming_e2e} proposed a CTC-based streamable E2E SLU framework which employs a unidirectional RNN to make multiple intent predictions. The NLU output is generated either directly from the audio signal or based on an intermediate ASR output. 
Namely, the advantage of semantic posterior that we employ, was not considered in making word-piece predictions.
To the best of our knowledge, there is no existing E2E SLU model, either streamable or non-streamable,  which takes both word-piece and slot tag in beam search decoding on joint multi-task sequence prediction for word-piece, slot tag, and intent label.
\section{Methodology}
\label{sec:methodology}

The inputs of the multi-task RNN-T are $D$-dimensional audio features of length $T$, $\mathcal{X} = (\mathbf{x}_1, \mathbf{x}_2, ..., \mathbf{x}_T)$, $\mathbf{x}_k \in \mathbb{R}^{D}$. The outputs are transcript tokens of length $U$, $\mathbf{y}^w = (y^w_1, y^w_2, ..., y^w_U)$, $y^w_u \in \mathcal{W}$, its corresponding slot tags (also of length $U$), $\mathbf{y}^s = (y^s_1, y^s_2, ..., y^s_U)$, $y^s_u \in \mathcal{S}$, and the intent $y^i \in \mathcal{I}$; here $\mathcal{W}$, $\mathcal{S}$, and $\mathcal{I}$ are the predefined set of token labels (or token vocabulary), slot tags, and intents. Both transcript tokens and slot tags are encoded as one-hot vectors. 

The model defines a conditional distribution of $p(\mathcal{W},\mathcal{S},\mathcal{I}|\mathcal{X})$, and we factorize it as follows (Fig.~\ref{fig:modelb}),
\begin{align}
    & p(\hat{\mathbf{y}}^w,\hat{\mathbf{y}}^s,y^i|\mathcal{X}) = \nonumber \\ 
    & \prod_{k=1}^{T+U} p(\hat{y}^w_k|\mathcal{X},t_k,y^w_0,...,y^w_{u_{k-1}},y^s_0,...,y^s_{u_{k-1}}) p(y^i|\hat{y}^w_k) \nonumber \\
    & \prod_{j=1}^{T+U} p(\hat{y}^s_j|\mathcal{X},t_j,y^s_0,...,y^s_{u_{j-1}},y^w_0,...,y^w_{u_{j-1}})
    \label{eq:cond_dist}
\end{align}
where $\hat{\mathbf{y}}^{w}=(\hat{y}^w_1,...,\hat{y}^w_{T+U}) \subset \{\mathcal{W} \cup \langle b^w \rangle\}^{T+U}$ , $\hat{\mathbf{y}}^{s}=(\hat{y}^s_1,...,\hat{y}^s_{T+U}) \subset \{\mathcal{S} \cup \langle b^s \rangle\}^{T+U}$ correspond to any possible alignment path with $T$ blank symbols and $U$ token/slot labels such that after removing all blank symbols, $b^w$ and $b^s$, in $\hat{\mathbf{y}}^{w}$ and $\hat{\mathbf{y}}^{s}$ correspondingly, it yields $\mathbf{y}^{w}$ and $\mathbf{y}^{s}$. $y^{w}_0$ and $y^{s}_0$ are the start of sentence and slot symbol respectively.

\subsection{Multi-task Semantic RNN-T} 
\begin{figure}[t]
\centering
\includegraphics[width=0.42\textwidth]{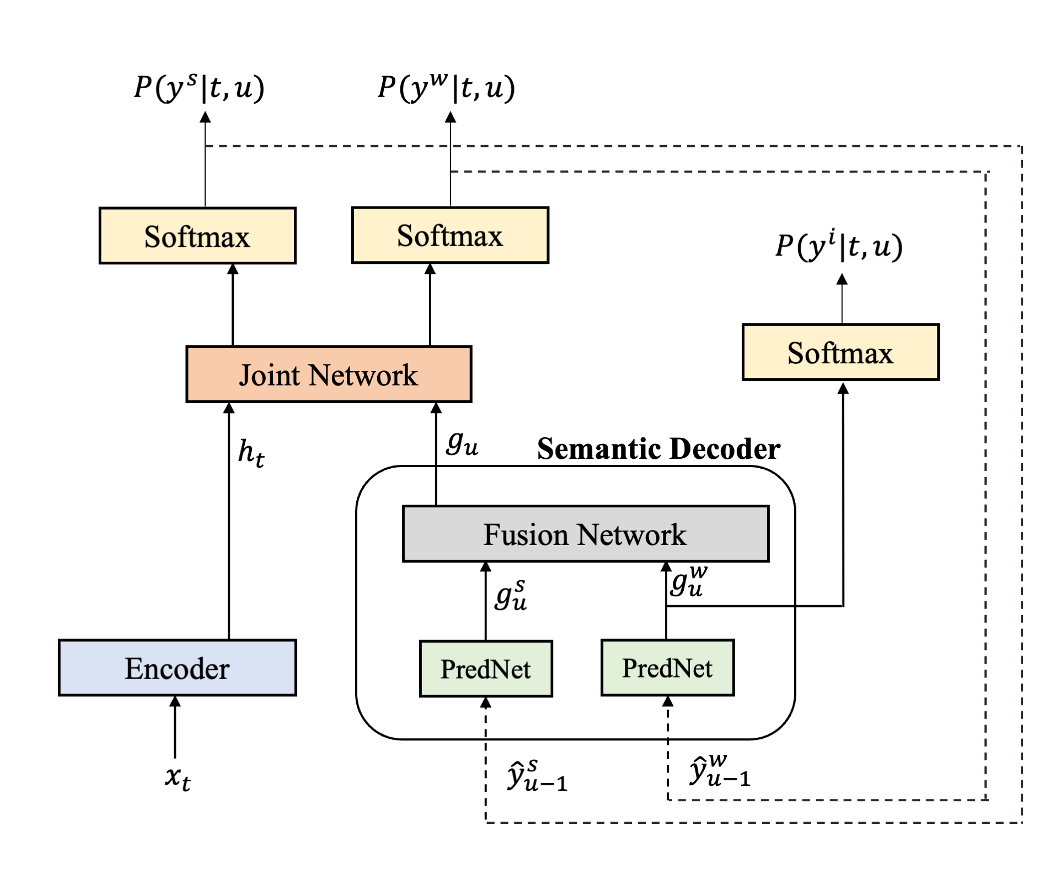}
\vspace{-8pt}
\caption{The proposed Multi-task Semantic RNN-T SLU model.} 
\label{fig:modelb}
\vspace{-10pt}
\end{figure}

The multi-task Semantic RNN-T architecture consists of three components, an audio encoder, a semantic decoder and a multiple-output joint network as shown in Fig.~\ref{fig:modelb}. The audio encoder is a unidirectional RNN \cite{lstm} that takes the audio features $\mathcal{X}  = (\mathbf{x}_1, \mathbf{x}_2, ..., \mathbf{x}_T)$ as inputs and generates the hidden representations, $\mathcal{H} = (\mathbf{h}_1, \mathbf{h}_2, ...,\mathbf{h}_T)$ auto-regressively. The semantic decoder takes in the word-pieces along with slot tags and outputs hidden label embeddings, $\mathcal{G}  = (\mathbf{g}_1, \mathbf{g}_2, ..., \mathbf{g}_U)$.
The encoder and semantic decoder output, $\mathbf{h}_t$ and $\mathbf{g}_u$, respectively, are then fed into a joint network to predict next word-piece, $y^w_{u+1}$, and slot tag, $y^s_{u+1}$.

The semantic decoder has two separate prediction networks for encoding word-pieces and slot tags, correspondingly, and a fusion layer is employed to aggregate the prediction network outputs.
Each of the prediction networks is a recurrent neural network consisting of an embedding layer, an output layer, and a recurrent hidden layer.
The outputs of the two prediction networks, $\mathbf{g}_u^w$ and $\mathbf{g}_u^s$, are then fused together producing the semantic decoder output, $\mathbf{g}_u$.
While we investigated both the addition, $\mathbf{g}_u = \mathbf{g}_u^w + \mathbf{g}_u^s$, and the concatenation with a projection, $\mathbf{g}_u = W([\mathbf{g}_u^w; \mathbf{g}_u^s]$) , as fusion methods, we did not observe a significant performance difference between them, so we report all results with the addition fusion method. 

Given the audio feature vector $\mathbf{h}_t$ and semantic decoder output $\mathbf{g}_u$, the multi-output joint network yields distributions for the word-piece and slot tag at the next time step $u+1$.
The joint network is composed of a feed-forward neural network and two separate classification layers to produce joint logits, also called lattice, for transcript tokens and slot tags,
$Z^{w} \in \mathbb{R}^{T \times U \times V^{w}}$ and $Z^s \in \mathbb{R}^{T\times U \times V^s}$, where $V^w$ and $V^s$ stand for the word piece size and slot value size, respectively. Each element of $Z^{w}$ and $Z^{s}$ represents the probability of the next word piece  $p(y^w_{u+1}|t, u)$,
and the next slot tag $p(y^s_{u+1}|t, y^s_u)$, correspondingly.
The intent classification layer is appended upon the prediction network for the word-piece.
The reason for separating intent prediction from the slot tag hypotheses is to reduce the effect of [Other] slot hypotheses (see Table~\ref{tab:semantc_labels}) in prior time steps on the intent prediction for the final state.

\vspace{-4pt}
\subsection{Semantic Beam Search}
To jointly decode the word piece and slot tag sequences at inference time, we propose a semantic beam search algorithm based on the search algorithm in \cite{rnnt} (only applied to the word-pieces) to find the top-n best output pairs of word pieces and slot tags. The motivation is to provide the decoder with the most possible candidate pairs of the fixed beam size in every time step and select the best aligned sequences through the decoding path. The semantic beam search parameters of each decoding are local word beam size ($B_{wp}$), local slot beam size ($B_{slot}$), local candidate pair beam size ($B_{local}$), and global candidate pair beam size ($B_{beam}$). $B_{wp}$ and $B_{slot}$ define the number of top possible word piece and slot candidates selected by the top log probability respectively. Given the $B_{wp} \times B_{slot}$ candidate pairs (combining the top possible word pieces and slot tags), $B_{local}$ of candidate pairs with the highest addition of log probabilities are then selected. Finally, among $B_{local} \times B_{beam}$ candidate pairs, we preserve $B_{beam}$ ones for the next decoding step.

\subsection{Loss Functions}
\label{subsec:losses}
\subsubsection{Word-Piece Loss}
The word-piece prediction is optimized with the RNN-T loss \cite{rnnt}, denoted as $L_{rnnt}(wp)$, which computes the alignment probability summation, $p(\hat{\mathbf{y}}^w|\mathcal{X})$ with a forward-backward algorithm.

\subsubsection{Intent Classification Loss}

The intent classification is optimized by minimizing the cross-entropy loss between the intent logits and the ground truth intent label, summed over a batch of utterances.
\begin{equation} 
\label{eq:intent_loss}
\begin{split}
        &L_{ce}(intent) = -\sum y^i \times \log(p(\hat{y}^i|t,u, y^w))
\end{split}
\end{equation}

\subsubsection{Slot Tagging Loss} 
Given the generated transcript tokens and slot tags from prediction networks (Fig.~\ref{fig:modelb}), this loss is designed to learn the alignment between the two sequences. In particular, we investigate two losses:
\begin{itemize}[leftmargin=12pt]
\item \emph{Cross Entropy Loss}: The cross-entropy loss is computed at each state of the slot lattice $Z^s$ and averaged over $T$ time steps for each decoder state.
\begin{equation}
    \begin{split}
         L_{ce}(slot) = -\sum_{u=1}^U\frac{1}{T}\sum_{t=0}^Ty^s \log(p(\hat{y}^s|t,u,y^w)) \
    \end{split}
\end{equation}
\item \emph{Aligned RNN-T Loss}: This loss consists of two terms as follows,
\begin{equation}
\label{joint}
\begin{split}
    L_{rnnt, align}(slot) = L_{rnnt}(slot) + L_{align}(slot)
\end{split}
\end{equation}
Similar to the word-piece loss, the first term, $L_{rnnt}(slot) $ is used to learn the alignment between the audio inputs and the slot tags with the standard RNN-T loss \cite{rnnt}. The second term $L_{align}(slot)$, is responsible for learning the alignment between the word pieces and their corresponding slot tags at each state, $L_{align}(slot) = -\log(p(\mathbf{y}^w, \mathbf{y}^s | \mathcal{X}))$. Based on a conditional independence assumption, we use $p(y^w_{u+1}, y^s_{u+1}|t, y^s_u, y^w_u) = p(y^w_{u+1}|t, y^w_u) \cdot p(y^s_{u+1}|t, y^s_u)$ at each state $(t, u+1)$ and are able to simply reapply the same transducer forward-backward algorithm on this combined lattice to efficiently compute $L_{align}(slot)$.
\end{itemize}

\section{Experimental Results}

% Table1: relative improvements of ASR and NLU metrics
\begin{table}[t]
\centering
\tabcolsep=0.1cm
\caption{Relative Improvements of ASR and NLU metrics (\%) for Multi-task Semantic RNN-T over the two-stage SLU~\cite{s2i}.}
\label{tab:mt_sem_rnnt}
\begin{tabular}{ccccc}
\toprule
Model  & WERR & SemERR & IRERR  & ICERR  \\ \midrule
Two-stage SLU~\cite{s2i} & 0 & 0 & 0 & 0   \\        
One-stage version of~\cite{s2i} & 0.6 & 1.2 & 0.1 & -5.8 \\
Multi-task Semantic RNN-T & \textbf{1.4} & \textbf{9.5} & \textbf{14.4} & \textbf{5.1} \\ \bottomrule
\end{tabular}
\vspace{-3mm}
\end{table}

% Table 2: Relative improvements
\begin{table*}[t]
\centering
\tabcolsep=0.1cm
\caption{Comparisons of different proposed multi-task losses}
\label{tab:different_losses}
\resizebox{0.8\linewidth}{!}{%
\begin{tabular}{cccccc}
\toprule
Model & Loss Type & WERR & SemERR & IRERR  & ICERR \\ \midrule
Two-stage SLU~\cite{s2i}   & - & 0 & 0 & 0 & 0               \\ %\midrule
\multirow{2}{*}{Multi-task Semantic RNN-T} & $ L_{rnnt}(wp) + L_{ce}(slot) + L_{ce}(intent) $   & \textbf{1.41} & \textbf{9.49} & \textbf{14.38} & \textbf{5.13}  \\
& $L_{rnnt}(wp) + L_{rnnt,align}(slot) + L_{ce}(intent) $ & -0.99  & \textbf{7.43}  & \textbf{12.04} & -1.26  \\ \bottomrule
\end{tabular}}%
\vspace{-2.5mm}
\end{table*}

% Table 4: relative value
\begin{table}[t]
\centering
\tabcolsep=0.1cm
\caption{Comparisons of different slot beam sizes, $B_{slot}$, in semantic beam search configuration, ($B_{wp}, B_{slot}, B_{local}, B_{beam}$), for Multi-task Sem-RNN-T}
\vspace{2.0mm}
\label{tab:beamsearch_seperate_intent}
\resizebox{0.8\linewidth}{!}{%
\begin{tabular}{ccccc}
\toprule
Semantic Beam Search & WERR & SemERR & IRERR  & ICERR \\ \midrule
(1,1,1,1)-Greedy Search & 0 & 0 & 0 & 0 \\
(10,1,10,8) & 8.5 & 0.6  & 0.6 & 1.1  \\
(10,2,10,8) & \textbf{8.6} & \textbf{11.9} & \textbf{9.5} & \textbf{2.9} \\ 
(10,4,10,8) & \textbf{8.6} & \textbf{11.9} & \textbf{9.5} & 2.8 \\ \bottomrule
\end{tabular}}%
\vspace{-4.5mm}
\end{table}

\begin{table}[h]
\centering
\tabcolsep=0.1cm
\caption{ASR and NLU performances (\%) on the public Fluent Speech Commands dataset}
\label{tab:fsc}
\begin{tabular}{c|c|c|c|c}
\toprule
Model  & Streaming & SemDec. & WER & IRER \\ \midrule
Two-stage SLU~\cite{s2i} & N & N & 0.61 & 0.85 \\
One-stage version of~\cite{s2i} & Y & N & 0.54 & 0.91 \\
MT Semantic RNN-T (Ours) & \textbf{Y} & \textbf{Y} & \textbf{0.55} & \textbf{0.84} \\ \bottomrule
\end{tabular}
\vspace{-3mm}
\end{table}

\subsection{Dataset} 
To evaluate the multi-task semantic RNN-T model, we use 1,300 hours of speech utterances from our in-house de-identified far-field SLU dataset, containing not only transcriptions but also slot tags and intents.
This is broken into training and test sets of 910 hours and 195 hours, respectively.
The device-directed far-field, speech data is captured using a smart speaker across multiple English locales (e.g. en-US, en-IN, etc.).
The input audio features fed into the network consist of 64-dimensional LFBE features, which are extracted every 10 ms with a window size of 25 ms from audio samples. The features of each frame are then stacked with the left two frames, followed by a downsampling of factor 3 to achieve a low frame rate, with 192 feature dimensions.
The subword tokenizer \cite{sennrich-etal-2016-neural, kudo2018subword} is used to create tokens from the transcriptions; we use 4000 word-pieces in total. There are 63 intent classes and 183 slot tags annotated in this dataset.
We also conducted experiments on the public SLU corpus, Fluent Speech Commands (FSC) dataset \cite{fluent}, which contains 15 hrs (23k utterances), 11 intents and 3 slots, including the $``Other"$ class.

\subsection{Baselines and Model Configurations}
We compare the proposed method with the state-of-the-art RNN-T based two-stage SLU model ~\cite{s2i}. We also compare to another baseline by extending \cite{s2i} to the one-stage version (as in Fig.~\ref{fig:high_level}) introduced in \cite{google}. The proposed model has the following configurations. The audio encoder is a 5-layer LSTM with 736 neurons and output size of 512-dimension. The word piece prediction network is of a 512-dim embedding layer followed by a 2-layer LSTM with 736 neurons and output size of 512-dim. The slot tag prediction network is of a 128-dim embedding layer and a 2-layer LSTM with 256 neurons and output size of 512-dim. The intent decoder consists of two 128-dim dense layers with $ReLU$ as an activation function. The joint network is a fully-connected feed-forward component with one hidden layer followed by a $tanh$ activation function.  Overall, the size of both the proposed model and the baselines sum to approximately 40 million parameters. For the FSC data set, given the small amount of transcribed data to train the ASR module of million parameters well, we followed \cite{s2i} by first pre-training the audio encoder, the word piece prediction network, and the joint network on 910 hrs Alexa data, before finetuning on the FSC dataset. 

\subsection{Metrics}
We use four metrics to evaluate the performance of an E2E SLU system.
(i) \textbf{Word Error Rate (WER)}: WER is a word-level metric used for evaluating the word-piece recognition performance. It calculates Levenshtein distance or edit distance that is the shortest distance required for transforming word-piece hypothesis to the ground truth by using insertion, deletion and substitution.
(ii) \textbf{Semantic Error Rate (SemER)}: The SemER metric jointly evaluates the performance of intent classification and slot filling or say NLU performance.
By comparing a word sequence reference and their accompanying slot tags, performance is calculated as: 
\begin{equation}
\label{eq:semer}
    SemER = \frac{\#Deletion+\#Insertion+\#Substitution}{\#Correct+\#Deletion+\#Substitution},
\end{equation}
where $Correct$ is when slot tag and slot value (words) are correctly identified, $Deletion$ is when a slot tag present in the reference is not the hypothesis, $Insertion$ is an extraneous slot tag included by hypothesis, and
$Substitution$ is when a slot tag from hypothesis is included but with the incorrect slot value. Intent classification errors are counted as substitution errors.
(iii) \textbf{Interpretation Error Rate (IRER)}: The IRER metric is an utterance-level metric for evaluating the joint intent classification and slot filling performance without partial credit. Namely, it is the fraction of utterances where either the intent or any of the slots are predicted incorrectly. 
(iv) \textbf{Intent Classification Error Rate (ICER)}: The ICER metric measures the error rate of intent classification, which is an utterance-level evaluation metric. 
The results of all experiments are presented as the relative error rate reductions (WERR/SemERR/IRERR/ICERR). For example, given model A's WER ($\text{WER}_A$) and a baseline B's WER ($\text{WER}_B$), the WERR of A over B is computed as
$\text{WERR} = (\text{WER}_B - \text{WER}_A)/\text{WER}_B.$

\vspace{-2pt}
\subsection{Results}
The results of multi-task semantic RNN-T over the baselines are shown in Table~\ref{tab:mt_sem_rnnt}. Jointly training ASR and NLU tasks with a semantic decoder shows consistent improvements across tasks and metrics by providing more contextual information. Of note is that the NLU metrics such as SemER and IRER are significantly improved while the improvement of WER is comparably small. We attribute this to the annotation bias of the slot tag distribution of the dataset: Around 60\% of the utterances have their slot tags mapped to the [Other] label. Therefore, the semantic information provided by slot tags for the word-piece generation may be limited.

In Table~\ref{tab:different_losses}, we compare different loss combinations as introduced in Sec~\ref{subsec:losses}. Using cross-entropy loss for slot tagging has demonstrated the best overall performances, and greatly improves the two-stage SLU model in terms of both NLU and ASR metrics. Imposing aligned RNN-T loss also significantly improves the NLU metrics such as SemER and IRER, but slightly degrades the WER and ICER. We believe this is because the loss is more sensitive to the misalignments between the word-piece and slot tags produced by the separate prediction networks. 

Finally, we validate the effectiveness of the semantic beam search algorithm. We fixed the best-performing parameters of $B_{wp}$, $B_{local}$, $B_{beam}$, and varied $B_{slot}$ to 1, 2, 4 and showed the results in Table~\ref{tab:beamsearch_seperate_intent}. As it can be seen, changing the parameters from a greedy search to the beam search, from (1,1,1,1) to (10,1,10,8), improves all metrics, while mainly improving WER. When further increasing the slot beam size from 1 to 2 or 4, the improvements over NLU metrics become significant, 
$11.9\%$ and $9.5\%$ in terms of SemERR and IRERR. Again, the improvement of WER from increasing slot beam search size is limited and may be attributed to the inherent bias of the slot tag annotations.

Table~\ref{tab:fsc} presents the results on FSC dataset~\cite{fluent}, where the streaming capability and the use of semantic information during decoding (Yes/No: Y/N) are also shown in the table. In our model (MT Semantic RNN-T), we used the additive fusion to obtain the semantic decoder output, with the semantic beam search size set by $B_{wp}$=10, $B_{slot}$=2, $B_{local}$=10, and $B_{beam}$=16. Note that due to the lower complexity and limited number of intents and slots in FSC, all the models in our experiments lead to $<1\%$ WER and IRER values. The proposed model improved the WER of 2-stage SLU by 9.8\% while improved IRER of 1-stage SLU by 7.7\% relatively. 
\section{Conclusion} 
In this paper, we proposed a multi-task semantic RNN-T architecture for streamable end-to-end spoken language understanding. The proposed semantic decoder and semantic beam search empower the model to consider more contextual information, both from the predicted word piece and slot tags in the history, and produce higher quality word-piece and slot tag hypothesis in the next time step. Moreover, the model with the proposed losses, both cross-entropy loss and the aligned RNN-T loss for slot tagging, outperformed the two-stage and one-stage E2E SLU models.

\section{Acknowledgement}
We would like to thank Gautam Tiwari and Anirudh Raju for valuable discussions on the work of this paper.

%\vfill\pagebreak

\bibliographystyle{IEEEbib}
\bibliography{reference.bib}

\end{document}